# Toward Generalized Detection of Synthetic Media: Limitations, Challenges, and the Path to Multimodal Solutions

Redwan Hussain[1]✉[0009-0004-3132-3432], Mizanur Rahman[1][0009-0008-2627-8664], Prithwiraj Bhattacharjee[1][0000−0001−9300−9351]

Department of Computer Science and Engineering
Leading University, Sylhet-3112, Bangladesh

{cse_182210012101026, cse_182210012101040, prithwiraj_cse}@lus.ac.bd

**Abstract-** Artificial intelligence (AI) in media has advanced rapidly over the last decade. The introduction of Generative Adversarial Networks (GANs) improved the quality of photorealistic image generation. Diffusion models later brought a new era of generative media. These advances made it difficult to separate real and synthetic content. The rise of deepfakes demonstrated how these tools could be misused to spread misinformation, political conspiracies, privacy violations, and fraud. For this reason, many detection models have been developed. They often use deep learning methods such as Convolutional Neural Networks (CNNs) and Vision Transformers (ViTs). These models search for visual, spatial, or temporal anomalies. However, such approaches often fail to generalize across unseen data and struggle with content from different models. In addition, existing approaches are ineffective in multimodal data and highly modified content. This study reviews twenty-four recent works on AI-generated media detection. Each study was examined individually to identify its contributions and weaknesses, respectively. The review then summarizes the common limitations and key challenges faced by current approaches. Based on this analysis, a research direction is suggested with a focus on multimodal deep learning models. Such models have the potential to provide more robust and generalized detection. It offers future researchers a clear starting point for building stronger defenses against harmful synthetic media.

## 1 Introduction

The evolution of Artificial Intelligence (AI) in media, focusing on image, audio, and video modalities, has significantly accelerated over the past decade. After the introduction of Generative Adversarial Network (GAN) by Goodfellow et al. in 2014 [1] the quality of photorealistic image generation has improved to an unprecedented level. It has become difficult to distinguish between real and AI-generated media. Early GAN based models gradually enhanced image quality until the diffusion-based model emerged and marked the beginning of an era of generative modeling.

The term 'Deepfake' first appeared in 2017 when certain Reddit users started exploiting GANs to create non-consensual adult content in which led to an increase in harmful AI-generated media. As a result, a range of problems appeared such as spread of misinformation, political conspiracies, privacy violations, fraudulent activities and the race between synthetic media generation and their detection. With the rise of diffusion-based models [2] since 2020, these tools can produce so highly realistic images and videos that are very difficult to distinguish from genuine content with the naked eye. Within the ongoing race of hundreds of models and their varying generation techniques demands the development of a generalized detection model to properly distinguish between real and synthetic data. Traditional detecting tools primarily rely on finding visual spatio-temporal anomalies, noise patterns, or the traces left behind by the generative models. These systems typically consist of deep learning (DL) architecture such as Convolutional Neural Network (CNNs) or Vision Transformers (ViT) to detect and classify properly. These models often struggle in becoming a generalized detector, failing to capture temporal anomalies properly, poor performance on unseen data. Furthermore, these detectors cannot perform properly on multimodal analysis on subtle or highly modified synthetic content. To combat these problems, developing a multimodal deep learning model needs a growing demand that can become a generalized content detector and fulfill all the limitation mentioned above.

The development of such a detector can bridge the gap between static detection methods and the dynamic evolution of synthetic media. It can also provide a robust protection on classifying harmful, altered media in our social media and daily life. A review of twenty-four recent studies on AI-generated content and deepfake detection methods was demonstrated. This study analyzed the key limitation and challenges faced by the researchers along with their models faced in the growing advancement of generative technology. The study includes:

- Identifies the overall key issues and challenges in AI-generated media detection that hinder research progress.
- Reviews recent studies individually and observes their methods, results, and weaknesses, and then summarizes the core limitations of the field based on these observations.
- Highlights the potential of multimodal approaches to improve detection where traditional methods are saturated.
- Provides clear directions for future research by offering a structured starting point for building robust and generalized detection systems.

## 2 Key Issues & Challenges

Detecting synthetic media especially, for video, is a highly complex task that faces several key challenges and limitations. As illustrated in figure-1, several of these issues have been showcased where some issues faced by current researchers and others are technical limitation of these models. Furthermore, we discuss all the possible challenges on developing a multimodal generalized detector. One major challenge is the lack of a diversified dataset that contains datasets from possibly all generative techniques. This allows the model to learn from a wide range of generators and increase the chances of accurately classify data from different models and become a generalized detector. However, the diversified large dataset brings a problem of computational limitation as Convolutional Neural Network (CNNs) models can take a lot of computational cost.

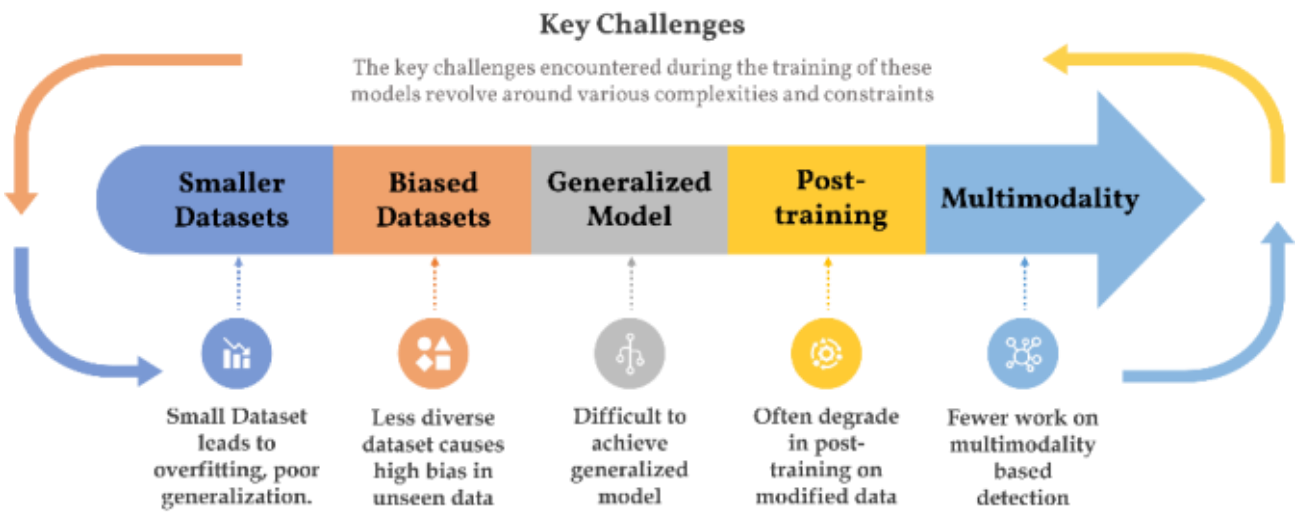

Fig 1: Key Issues & Challenges

Another key challenge for CNN based models is the capturing of subtle temporal anomalies. While CNN models are very good at capturing spatial patterns and features within still images, but is not inherently designed for capturing temporal dynamics. The reasons for the failure of CNN models are overfitting in training data, not being able classify data generated by different algorithms and prioritizing spatial feature extraction over temporal modeling due to their fundamental architecture design. To address this issue, we can use Vision Transformer (ViT) for temporal learning. Unlike traditional CNNs that analyze each frame in isolation (better in image detection) ViT based model can be configured by converting a video into a sequence of frames. Then again ViTs are very data-hungry models due to not having any inductive biases like CNNs. It makes them treat every frame as a starting point and learns locality and translation variance from the raw. This makes CNNs learn everything solely from the training data rises the issue of computational cost for a larger dataset. To resolve this issue, some researchers leverage pre-trained ViTs combined with a few-shot learning approach [3] to minimize computational cost. But this has its own problems such as domain-specific mismatch, lack of generalization on unseen data etc. which can lead to an overfit as few-shot approaches uses very small amount of data in training that may create a post-training bottleneck on unseen data and also pre-trained ViTs may not be entirely relevant to the new specialized domain as they are trained in massive general type dataset. Ultimately any detection model is constrained by both its limitations and the challenge of acquiring a proper diversified dataset hindering a need for developing a better model built for generalized detection.

## 3 Literature Review

A total of twenty-four paper were selected for demonstration. Table 1 illustrate the comparative analysis of these papers with state-of-the-art techniques and multimodality mentioned and Figure 2 shows a generalized diagram of all these problems summarized. D. Tan et al. [4] conducted a thorough review on deep-learning based multimodal forgery detection for video and audio by utilizing fusion techniques and finds high performance metrics (ACC: 95.11%, AUC: 99.50%, AP: 98.50%, EER: 1.32%) across multiple benchmark datasets (UADFV, FF++, and Celeb-DF v2). However, a cross model generalization limits the effectiveness in practical scenarios. C. Zheng et al. [5] advanced AI-generated video detection by introducing a model D3: that analyzes content using temporal dynamics using L2 distances of second-order derivatives. With the limitation of the inability to handle low-resolution video content. H. Wen et al. [6] developed BusterX++, a unified multimodal detector that analyzes image and video content using pre-trained Large Language Models (LLMs) with hybrid reasoning and post-training reinforcement. The study's findings indicate high performance (ACC: 93.9%, F1: 93.7%) on So-Fake-Set and GenBuster-200k. A notable limitation is the post-training bottleneck that can reduce performance over time. K. Veeramachaneni et al. [7] proposed a detection framework based on features extracted from pre-trained large vision models. It reduces the use of large dataset and extensive training. F1 score is 85.1% in the VID-AID dataset was shown with a limitation on being based heavily on spatial and limited temporal features. C. Internò et al. [8] proposed ReStraV, which uses temporal and perceptual cues captured via a self-supervised pre-trained model (DINOv2). The study's findings indicate strong performance (ACC: 90.90–97.06%, mAP: 98.81–99.12%) on VidProM, GenVidBench, and Physics-IQ datasets, with a possibility of poor performance on video analysis due to being solely based in image detection.

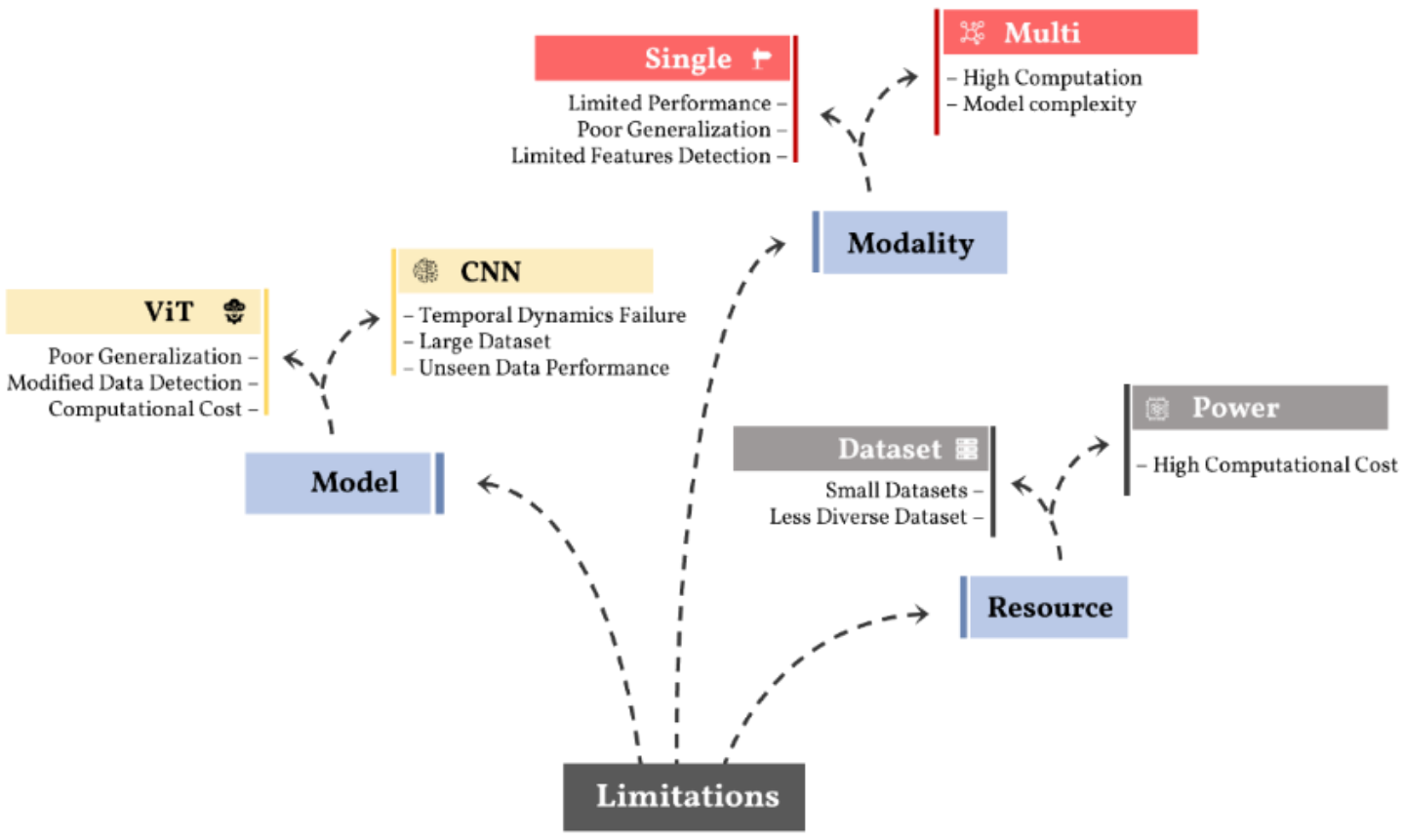

Fig 2: A Generalized Limitaion

P. James [9] proposes a framework by combining facial-visual features and audio features for deepfake detection by utilizing pre-trained Xception and DenseNet models for visual analysis and MFCC/Mel-spectrogram for audio analysis. A significant limitation is its dependence on pretrained models and specific feature selection can limit generalization to diverse datasets. R. Corvi et al. [10] contributed a forensic-oriented AI-generated video detection framework that uses wavelet transforms to analyze frequency-level traces in video frames. On the Panda70M dataset (bACC: 94.3%) the study demonstrates a strong performance with a limitation of training the model on a single generator making not a very ideal detector. J. Battocchio et al. [11] advanced AI-generated video detection by using a ViT backbone with temporal integration to capture both spatial and temporal artifacts. The study demonstrates strong performance on datasets (Youtube-8M, Sports-1M, FloreView, and Socrates) and better performance in accuracy metrics (TPR: 95%, TNR: 93%, AUC: 67–96%, ACC: 68–94%). L. H. Singh et al. [12] made a comprehensive multimodal dataset for audio video deepfake detection and benchmarking multiple detection methods. It enables fine grained evaluations and promotes multimodal detection research. Y. Zou et al. [13] analyze existing non-MLLM and MLLM-based detection techniques and evaluate their performance across multiple public datasets.

The study highlights gaps in multimodal integration and localization. I. Balafrej and M. Dahmane [14] advanced detection by introducing a steganography-inspired CNN with multi-level face detection. It also utilizes a confidence-based frame aggregation and enables efficient video-level predictions and cross dataset generalization. Its limitation is, the detection accuracy depends heavily on face detection quality. R. Kundu et al. [15] introduce a universal transformer-based detector (UNITE) for AI-generated content. Its capable of distinguishing fully synthetic videos as well as partial manipulations using comprehensive frame analysis. The model achieves 97–99% accuracy on standard datasets. J. Bai et al. [16] contributed by developing a two-branch spatial-temporal model to detect AI-generated videos by capturing pixel-level anomalies and motion inconsistencies. However, a potential overfitting with certain fusion methods can be seen in detecting subtle image-to-video (I2V) manipulations. Y. Wang et al. [17] developed an audio-visual multimodal detection approach by integrating audio-visual local-global interactions that capture both temporal audio cues and spatial-visual features. They demonstrated strong performance (ACC: ~94–97%, AUC: ~0.95–0.97) across benchmark datasets with a reliance on high-quality audio that might limit practicality for a real-world scenario. X. Liu et al. [18] explored evaluation methodologies for AI-generated video detection and introduced a multimodal benchmarking approach and assessment metrics such as GPT-4V eval, LLMScore, VIEScore, TIFA, TIGEr, ViLBERTScore, and COSMic. X. Song et al. [19] developed a multimodal diffusion forgery detector by integrating IAFA, LLaVA, CLIP, and VQ-VAE while demonstrating strong accuracy (AUC: 92.0%). Oorloff et al. [20] proposed AVFF, a multimodal framework combining speech and facial embeddings for video deepfake detection. This model achieves 98.6% ACC and 99.1% AUC on FakeAVCeleb. D. S. Vahdati et al. [21] investigate domain gaps between synthetic images and videos and report that image-trained models achieve limited generalization on video benchmarks (AUC: ≈74%). While few-shot learning improves robustness under compression it brings the problem of post-training bottleneck on unseen data. L. Ma et al. [22] propose DeCof, a transformer-based framework utilizing frame-level inconsistencies to detect AI-generated videos. Their model achieves 98.10–99.72% accuracy and 85.87–96.43% AUC on the GVF dataset. S. Muppalla et al. [23] introduced a multimodal deepfake detection framework using Cross-Modal Alignment and Distillation (CAD) to fuse audio-visual embeddings. Thus, improving benchmark performance but facing limitations with asynchronous or corrupted modalities. It is tested on FakeAVCeleb, TMC dataset providing an improved accuracy on multiple performance metrics. V. Vora et al. [24] propose a multimodal framework combining BERT for text analysis and CNN for image detection. Their model achieves ACC of 80.67% and an AUC of 83.67% on CIFAKE and ML Olympiad datasets. D. Salvi et al. [25] develop an end-to-end multimodal detection approach by leveraging CNN-LSTM networks. Around ACC: ~96% and AUC: ~97% on FF++, FakeAVCeleb, and DFDC subsets. But its effectiveness can be compromised by noisy or low-quality audio-visual data. Khalid et al. [26] integrates facial-visual cues with a pre-trained CNN model and audio features including MFCCs and emotional representations which leads to improvement in multimodal deepfake detection. Y. Zhou and S. Lim [27] proposed a multi-stream audio-visual detection model combining spatial-temporal video cues, speech waveform artifacts, and lip–speech synchronization features. It leverages R (2+1) D-18 CNN for video and a 1D CNN for audio, fused with an attention-based mechanism. With an effective result on FF++ and DFDC.

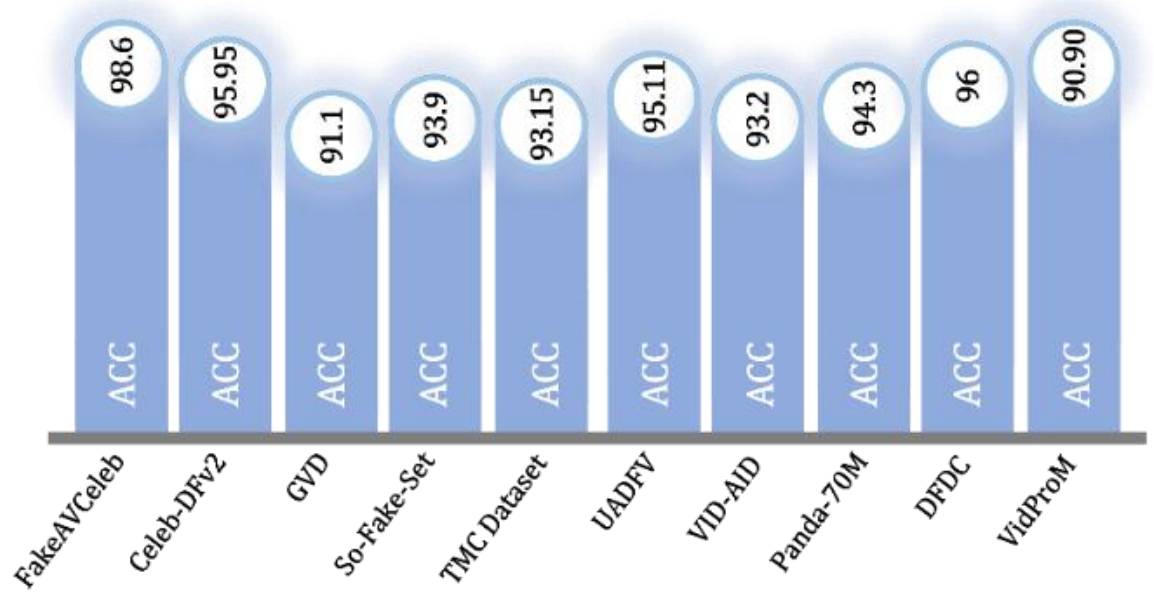

Fig 3: Accuracy (ACC) Score on Multiple Dataset Corresponding to Model Used

Table 1: Comparative analysis of existing detection model and their performance with limitation

| Source | Problem statement | Model, modality | Dataset used | Performance matrix | Contribution and solution | Limitation |
|---|---|---|---|---|---|---|
| Tan et al. [4] | Increase of deepfake technologies needs advance detection for cross-domain generalization. | Multi [audio-video] | UADFV, FF++, Celeb-DF v2 | ACC: 95.11% AUC: 99.50%, AP: 98.50% | Multi-source information and better fusion can capture better forgery traces. | Does not talk about any limitation of any models. |
| Zheng et al. [5] | Existing techniques struggle to capture temporal artifacts in high-fidelity synthetic videos. | XCLIP-B/16. Single [video] | Youku-mPLUG, Pika | mAP: 98.46 AUC: 93.45% | D3: calculates the L2 distance of the spatial features, trained on a small dataset. | Heavily relies on visual features encoders, can not handle low-quality videos. |
| Wen et al. [6] | Cross modal synthetic content creates a need of unified detection framework capable enhanced reasoning credit and eliminates cold start issues. | Qwen2.5 Instruct, Sophia Thinking, LoRA. Multi [image-video] | So-Fake-Set, GenBuster-200k. | ACC: 93.9%, F1: 93.7%. | BusterX++ a unified MLLM with reinforced learning in post-training with hybrid reasoning. | Advance generation techniques leave gap for continual adaptation of detecting techniques. |
| Veeramachaneni et al. [7] | Detection technique focused on the human faces creates need of a framework capable of handling non-face deepfakes. | VideoMAE, SigLIP. Single [video] | VID-AID | F1: 85.1% | Model built on feature extracted from pre-trained visual models and classify based on feature distance. | Relies heavily on pre-trained vision models. |
| Interno et al. [8] | Pixel-based detector fails detecting temporal inconsistency and needs temporal and perceptual patterns. | DINOv2. Single [image] | VidProM, GenVidBench, Physics-IQ | ACC: 90.90-97.06%, mAP: 98.81-99.12% | ReStraV a perceptual straightening approach to accurately classify. | Model also relies heavily on pre-trained visual models. |
| James et al. [9] | Existing detection models are unimodal and fails to utilize complementary information from cross-modalities. | Xception, DenseNet, (MFCC). Multi [audio-video] | FakeAVCeleb | ACC: 98.6%, AUC: 98.8% | A robust multimodal detection framework using pretrained audio and visual models | Limited by class imbalance and relies on pre-trained models and selected features. |
| Corvi et al. [10] | Video forensic detectors detect realistic content but lack in generalization for focusing only high-level artifacts. | Pyramid Flow. Single [video] | Panda70M | bACC: 94.3% | Forensic oriented data augmentation technique based on wavelet transform of video frames. | Not designed to capture temporal artifacts, trained by a single model video. |
| Battocchio et al. [11] | AI-based multimedia concerns of widespread misinformation. | Vision transformer Single [video] | Youtube-8m, Sports-1m, Vision, FloreView Socrates | TPR: 95%, TNR: 93%, AUC: 67-96%, ACC: 68-94% | ViT + few-shot learning is used in the model, adapts well in unseen data. | Generalization to unseen data is a limited due to few-shots learning. |
| Singh et al. [12] | Current deepfake datasets lack accurately synchronized audio-video pairs and detailed labels. | Benchmarked 11 different methods. Multi [audio-video] | FakeAVCeleb | ACC and AUC value change in different methods. | First large-scale, racially balanced multimodal dataset with synchronized audio-video. | limited size and coverage, with future plans to improve dataset maintenance. |
| Zou et al. [13] | There is lack of generalized video detection technique as per the evolution of AI generated content. | Non-MLLM methods, MLLM methods | Multiple public datasets have been used | Multiple performance matrix has been used | Hybrid approaches on combining multiple single modalities with domain specific detector. | Significant gap in use of multimodality in generalized detector for synthetic media. |
| Balafrej and Dahmane [14] | Deepfake detectors are resource-heavy, slow, process too many frames, making them impractical for deployment. | CNN, SRNet-based CNN. Single [Image] | Celeb-DFv2, DFDC | Celeb-DFv2: ACC:95.95 % AUC:99.13 % DFDC: ACC:81.52 % AUC:88.11 % | Stepwise face detection approach, confidence-based frame aggregation for pixel-level artifacts. | Lower accuracy on challenging dataset, and dependence on face detection quality. |
| Kundu et al. [15] | Existing detection model fails to detect content when no face available and the background is generated by AI. | SigLIP-So400m, MHSA transformer Single [image] | DeMamba, FF++, GTA-V | Attention diversity loss + cross entropy loss: 97-99% | UNITE a transformer-based model with multiple attention heads. | -- |
| Bai et al. [16] | Existing methods lack effective generalized detection techniques for AI-generated videos addressing spatial and temporal inconsistencies. | AIGVDet, ResNet50 Multi [image-video] | GVD (Generated Video Dataset) | T2V ACC: avg 91.1%, AUC: avg 96.9% I2V ACC: avg 89.7%, AUC: avg 96.1% | AIGVDet a model combining spatial domain detection on RGB frames with temporal domain detection based on optical flow. | Method depends on accurate optical flow, struggles with subtle I2V videos, and is sensitive to preprocessing. |
| Wang et al. [17] | Deepfake videos are becoming more realistic, and existing models often fail to generalize or remain robust. | AV-LGNN, EResNet, AV-LGI. Multi [audio-image] | FF++, Celeb-DF, DFDC, FakeAVCeleb, and Forgery Net | ACC: around 94.2% - 96.8% AUC: around 0.95 - 0.97 | Novel audio-visual interaction framework with attention to weigh local-global audio cues. | High computational cost, depending audio quality. |
| Liu et al. [18] | Performance of detection methods and meet semantic goals with given human perception. | -- | EvalCrafter, Chivileva, T2VBench | GPT-4V eval, LLMScore, VIEScore, TIFA, TIGEr, ViLBERTScore, COSMic | Highlights importance of AI generated video evaluation as a distinct research area. | -- |
| Song et al. [19] | Existing algorithms focuses on facial forgeries and often fail to identify diffusion generated content. | LLaVA, CLIP, VQ-VAE, IAFA. Multi [text-video]. | DVF | AUC: 92.0% | MM-Det a multi modal forgery detection with reasoning and spatio-temporal analysis. | Depends on LLM and may face generalization difficulties on unseen data. |
| Oorloff et al. [20] | Most deepfake detectors are uni-modal, rely on supervised training of dissonance patterns. | AVFF Multi [audio-video]. | FakeAVCeleb dataset. | ACC: 98.6%, AUC: 99.1% | Combined self-supervised learning, cross-modal strategy for accuracy. | Depends on audio-visual synchronization, underfit on async videos |
| Vahdati et al. [21] | Synthetic video generator leaves different traces than image generators. | ResNet-50, ResNet-34, Xception, DenseNet Single [image] | COCO, LSUN, MIT, Video-ACID. | AUC: avg 74%, RER: 76-96% | Few-shot learning detects accurately even after compression of traces left from different models. | Modified synthetic video can deceive the model to capture the traces. |
| Ma et al. [22] | AI generated video have raised a potential security concern. | CLIP:ViT Single [video] | GVF (proposed dataset) | ACC: 98.10-99.72%, AUC: 85.87-96.43% | DeCof, detects via temporal artifact, and distinguish on videos properly. | Evaluate based on temporal artifact only. |
| Muppalla et al. [23] | Current deepfake detectors are mostly single-modal and fail to capture audio-visual evidence. | CAD. Multi [audio-video] | FakeAVCeleb, TMC Dataset | ACC: 99.20 ~ 93.15%, AUC: 99.30 ~ 93.67% | Deepfake detection framework with specific signals and cross-modal alignments. | Model may struggle with asynchronous or corrupted modalities. |
| Vora et al. [24] | AI generated content will become harder to detect using generalized model as the technologies will get better day by day. | BERT, CNN. Multi [text-image] | CIFAKE, ML Olympiad Competition. | AUC: 83.67%, ACC: 80.67%, | BERT+CNN multimodal by leveraging NLP for text and CNN for image detection. | Smaller, unvaried dataset, working with audio as future research scopes. |
| Salvi et al. [25] | Deepfakes manipulate audio and video, but existing detectors often overfit to unimodal patterns due scarce multimodal datasets. | CNN, LSTM, MFCC, SVM. Multi [audio-image]. | FF++, ASVspoof 2019 Evaluation FakeAVCeleb, DFDC (subset) | ACC: Around 96% Multimodal AUC: around 97% | Speech and spectral audio features with spatial and temporal visual cues, fused via temporal modeling. | Performance depends on the quality of unimodal training data, can degrade with poor audio-visual |
| Khalid et al. [26] | Current deepfake detectors are mostly single-modal and fail to effectively capture audio-visual evidence. | MesoInception-4, Meso-4, Xception, Multi [audio-image] | FakeAVCeleb | ACC: nearly 100% (audio) 61%-90% (video) AUC: 97.96% | Uses pretrained CNNs for visual and audio features, applies late fusion and ensemble voting. | Weaker visual detection, emotion-based features required |
| Zhou and Lim [27] | Existing deepfake detection methods focus only on visual manipulations, overlooking audio deepfakes and cross-modal inconsistencies. | R (2+1) D-18 CNN (visual), 1D CNN (audio). Multi [audio-image] | FF++, DFDC | FF++ - ACC: 97.02%, AUC: 99.65 DFDC – ACC: 91.01%, AUC: 96.32% | A two-plus-one-stream model that jointly trains audio, video, and sync-streams, uses attention-based feature fusion. | Lack of a high-quality, large-scale dataset was mentioned |

## 4 Future Research Direction

Based on the literature review of these twenty-four papers, some research directions appear to address the gaps mentioned above. The primary direction is to build a proper generalized synthetic media detector that can classify properly across data generated from various models and generation techniques. The study also addresses the challenges of dataset limitation and resource limitation bringing a research scope of building a proper diversified benchmarking dataset that includes datasets generated from all kinds of generative techniques possible. It also mentions the gaps of multimodal detectors such as AI generated content detection based on not only spatio-temporal anomalies but also audio-video fusion along with image and video fusion. A proposed research scope is also given as how to leverage a pre-trained ViT using a few-shot learning algorithm that does not lack of proper generalization capabilities in a post-training scenario on unseen data variations. More robust research (fig: 4) can be done on building a CNN and LSTM model that can classify via audio-video fusion thus making a multimodal audio-video detection model. The study shades a proper light on the gap of using multimodality on detecting synthetic media among different generation techniques. The rise in the generation techniques is increasing rapidly against the improvement of their detection models. This makes an emergence need for tackling the harmful side of these content across our daily life by designing a proper generalized detection model on synthetic media.

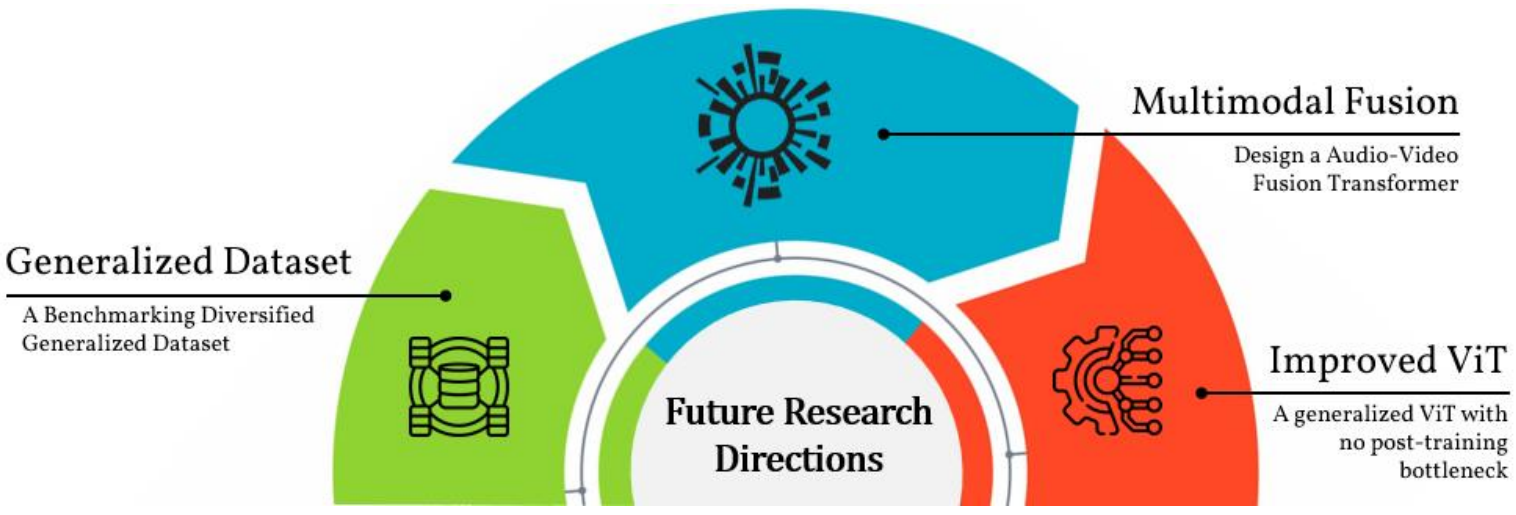

Fig 4: Research Scopes in Detecting Synthetic Media.

## 5 Conclusion

This study reviewed twenty-four recent works on AI-generated media detection. By analyzing each study, it identified the key challenges and limitations in the field, which include poor generalization on unseen data, difficulty handling content from different generative models, and limited performance on subtle or multimodal manipulations. The review also highlighted that traditional detection methods are reaching their limits. In contrast, multimodal approaches offer a promising path forward. Based on these observations, a clear research direction is suggested. The path is the development of generalized multimodal detection systems that can adapt to diverse and evolving synthetic media. These insights provide a structured foundation for future research. It can guide researchers in designing more robust and effective solutions to mitigate the risks of harmful AI-generated content.